\documentclass{article}

\usepackage[final]{neurips_2023}
\usepackage{hyperref}

\usepackage{natbib}
\setcitestyle{numbers,square}

\usepackage[utf8]{inputenc} 
\usepackage[T1]{fontenc}    
\usepackage{hyperref}       
\usepackage{url}            
\usepackage{booktabs}       
\usepackage{amsfonts}       
\usepackage{nicefrac}       
\usepackage{microtype}      
\usepackage{xcolor}         
\usepackage{multirow}
\usepackage{graphicx}
\usepackage{CJKutf8}

\title{WanJuan: A Comprehensive Multimodal Dataset for Advancing English and Chinese Large Models}

\author{%
  Conghui He \\
  Shanghai AI Laboratory \\
  \texttt{heconghui@pjlab.org.cn} \\
  \And
  Zhenjiang Jin \\
  Shanghai AI Laboratory \\
  \texttt{jinzhenjiang@pjlab.org.cn} \\
  \And
  Chao Xu \\
  Shanghai AI Laboratory \\
  \texttt{xuchao@pjlab.org.cn} \\
  \And
  Jiantao Qiu \\
  Shanghai AI Laboratory \\
  \texttt{qiujiantao@pjlab.org.cn} \\
  \And
  Bin Wang \\
  Shanghai AI Laboratory \\
  \texttt{wangbin@pjlab.org.cn} \\
  \And
  Wei Li \\
  Shanghai AI Laboratory \\
  \texttt{liwei@pjlab.org.cn} \\
  \And
  Hang Yan \\
  Shanghai AI Laboratory \\
  \texttt{yanhang@pjlab.org.cn} \\
  \And
  Jiaqi Wang \\
  Shanghai AI Laboratory \\
  \texttt{wangjiaqi@pjlab.org.cn} \\
  \And
  Dahua Lin \\
  Shanghai AI Laboratory \\
  \texttt{lindahua@pjlab.org.cn} \\
}

\begin{document}

\maketitle

\begin{abstract}

The rise in popularity of ChatGPT and GPT-4 has significantly accelerated the development of large models, leading to the creation of numerous impressive large language models(LLMs) and multimodal large language models (MLLMs). These cutting-edge models owe their remarkable performance to high-quality data. However, the details of the training data used in leading paradigms are often kept confidential. This lack of transparency, coupled with the scarcity of open-source data, impedes further developments within the community. As a response, this paper presents "Wan Juan", a large-scale multimodal dataset composed of both Chinese and English data, collected from a wide range of web sources. The dataset incorporates text, image-text, and video modalities, with a total volume exceeding 2TB. It was utilized in the training of InternLM, a model that demonstrated significant advantages in multi-dimensional evaluations when compared to models of a similar scale. All data can be accessed at \href{https://opendatalab.org.cn/WanJuan1.0}{https://opendatalab.org.cn/WanJuan1.0}. 
  
\end{abstract}

\section{Introduction}

Before the advent of large language models~\cite{devlin2018bert, radford2018improving, radford2019language, brown2020language, chatgpt, openai2023gpt4, team2023internlm} and multimodal models~\cite{radford2021learning, li2022blip, li2023blip, dai2023instructblip}, the NLP and CV fields mostly used small amounts of high-quality manually annotated data to train models and conduct research on various specific tasks. Research at this stage focused more on high-quality domain-specific data, improving model performance through enhancing model structures, leading to a large number of excellent network structures. However, with the introduction of large models like Bert and BLIP, pretraining with a large amount of unsupervised internet data can give models good generalization capabilities. Based on large-scale pretraining, using a small amount of SFT fine-tuning and RLHF fine-tuning, some tasks even exceed the average human level. These studies all reflect the importance of large-scale pretraining datasets.

More and more research teams are releasing large-scale datasets, such as C4~\cite{raffel2020exploring}, Pile~\cite{gao2020pile} in the NLP field, and LAION400M~\cite{schuhmann2021laion}, LAION-5B~\cite{schuhmann2022laion}, CC3M~\cite{sharma2018conceptual}, CC12M~\cite{changpinyo2021conceptual}, MMC4~\cite{zhu2023multimodal} in the multimodal research field. These datasets are far more significant in volume than supervised data, contain relatively affluent information, and effectively promote the development of LLMs and MLLMs.

In this work, we consider the diversity of dataset modalities, content diversity, content safety, and content quality. Based on these basic principles, we have collected, processed, and screened text, image-text, and video data from the Internet. The text data covers multiple fields, including technology, literature, media, education, and law; the image-text data covers a variety of fields, such as news events, people, natural landscapes, and social life; the video data covers military, arts, sports, nature, real world, knowledge, film art, media, food, history, science, and education, etc. These pretraining data have significantly improved the knowledge content, logical reasoning, and generalization abilities of the training model. Specifically, our contributions are as follows:

\begin{itemize}
    \item We build a large-scale training corpus \textbf{WanJuan} that includes multiple modalities:  the text data includes more than 600 million documents, with a data storage volume exceeding 1TB; the image-text data is processed into  documents, with a total of more than 22 million documents and a data size exceeding 200GB (images are provided via URL links); the video files total over 1000, with a data size exceeding 900GB.

    \item In the construction process of the WanJuan dataset, we ensure data safety, high quality, and value alignment (filtering out pornography, violence, and bias) through algorithmic processing and manual verification.

    \item We provide a unified JSON format processing, dataset download tool, and supporting documentation to facilitate users in quickly applying large model training.

\end{itemize}

\section{Dataset Statistics}

The \textbf{WanJuan} dataset includes multimodal data such as text, image-text, and video, all in Chinese or English, covering numerous fields and boasting high quality. Each modality has been carefully selected and processed to ensure the diversity and comprehensiveness of the dataset. Specifically, the composition of the dataset is as follows:

\subsection{Text Data}

\begin{table*}[t]
\centering
\begin{tabular}{llccc}
\toprule
Data Type & Language/Source & Weight(\%) & Number of Files & Size(GB) \\
\midrule
\multirow{8}{*}{Text Data}  

 & EN/WebText   & 61.4  & 383M & 434.7 \\
 & CN/WebText   & 35.3  & 220M & 505.1 \\
 & CN/Law       & 1.3   & 8M   & 35.3\\
 & CN/ChinaNews & 1.1   & 7M   & 20.0 \\
 & CN/Exam      & 0.6   & 4M   & 4.3 \\
 & CN/Patent    & 0.2   & 1M   & 17.2 \\
 & CN/TextBook  & 0.07  & 454K & 2.2 \\
 & CN/Wiki      & 0.01  & 92K  & 0.1 \\ 
 & Total           & 100   & 624M & 1019.7  \\ 
\bottomrule
\end{tabular}
\caption{Text Data Statistics.}
\label{tab:text_data}
\end{table*}

\begin{figure*}[t]
\begin{center}
	\includegraphics[width=1 \linewidth]{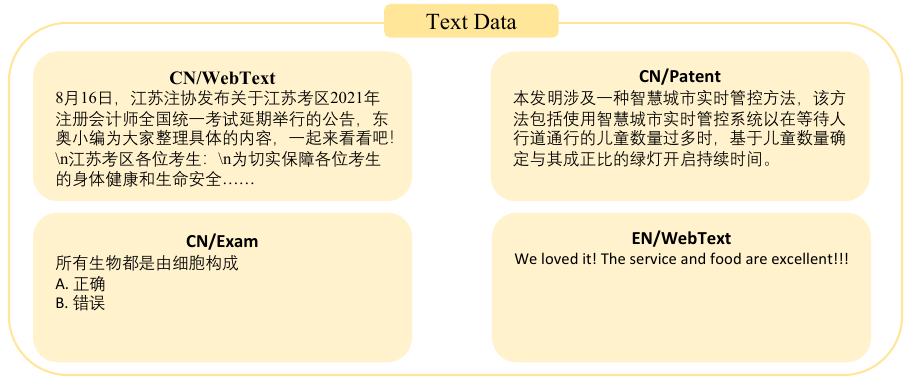}
\end{center}
		\caption{ Examples of Text Data }
\label{fig:fig1-text}
\end{figure*}

The text data comes from different sources such as web pages, encyclopedias, books, patents, textbooks, and exam questions. Through finely designed rules and algorithms, we filter and process the original data, remove invalid content, and ensure the content's safety and high information content. The specific composition of the text data is shown in Table~\ref{tab:text_data}, which includes more than 600M documents with a storage volume exceeding 1TB. Figure ~\ref{fig:fig1-text} provides some examples of the text data. This extensive collection of text data provides a rich resource for training language models and studying various NLP tasks.

\subsection{Interleaved Image-Text  Data}

\begin{figure*}[t]
\begin{center}
	\includegraphics[width=1 \linewidth]{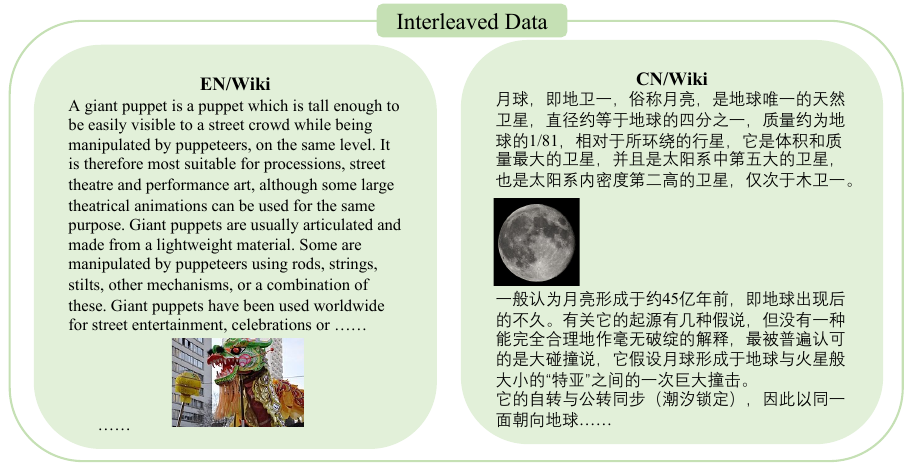}
\end{center}
		\caption{ Examples of Image-Text Interleaved Data }
\label{fig:fig2-image-text}
\end{figure*}

The image-text multimodal data originates from public web pages. After processing, it forms interleaved image-text documents. The specific composition is shown in Table 2, which includes over 22 million documents and a data size exceeding 200GB (excluding images). Figure~\ref{fig:fig2-image-text} presents some examples from the interleaved image-text dataset. The interleaved image-text data provides a valuable resource for studying the interaction between text and images, which is crucial for many multimodal tasks.

\begin{table*}[t]
\centering
\begin{tabular}{llccc}
\toprule
Data Type & Language/Source & Weight(\%) & Number of Files & Size(GB) \\
\midrule
\multirow{4}{*}{Image-Text Data} 
 
 & EN/Wiki                      & 37.7 & 9M  & 75.8 \\
 & CN/Authoritative Media News  & 5.3  & 2M  & 10.7 \\ 
 & CN/Self-Media News           & 53.4 & 10M & 107.4 \\ 
 & CN/Wiki                      & 3.6  &882K & 7.2 \\ 
 & Total                        & 100  & 22M & 201.1 \\ 
\bottomrule
\end{tabular}
\caption{Image-Text Interleaved Data Statistics.}
\label{tab:image_text_data}
\end{table*}

\subsection{Video Data}
\begin{figure*}[ht]
\begin{center}
	\includegraphics[width=1 \linewidth]{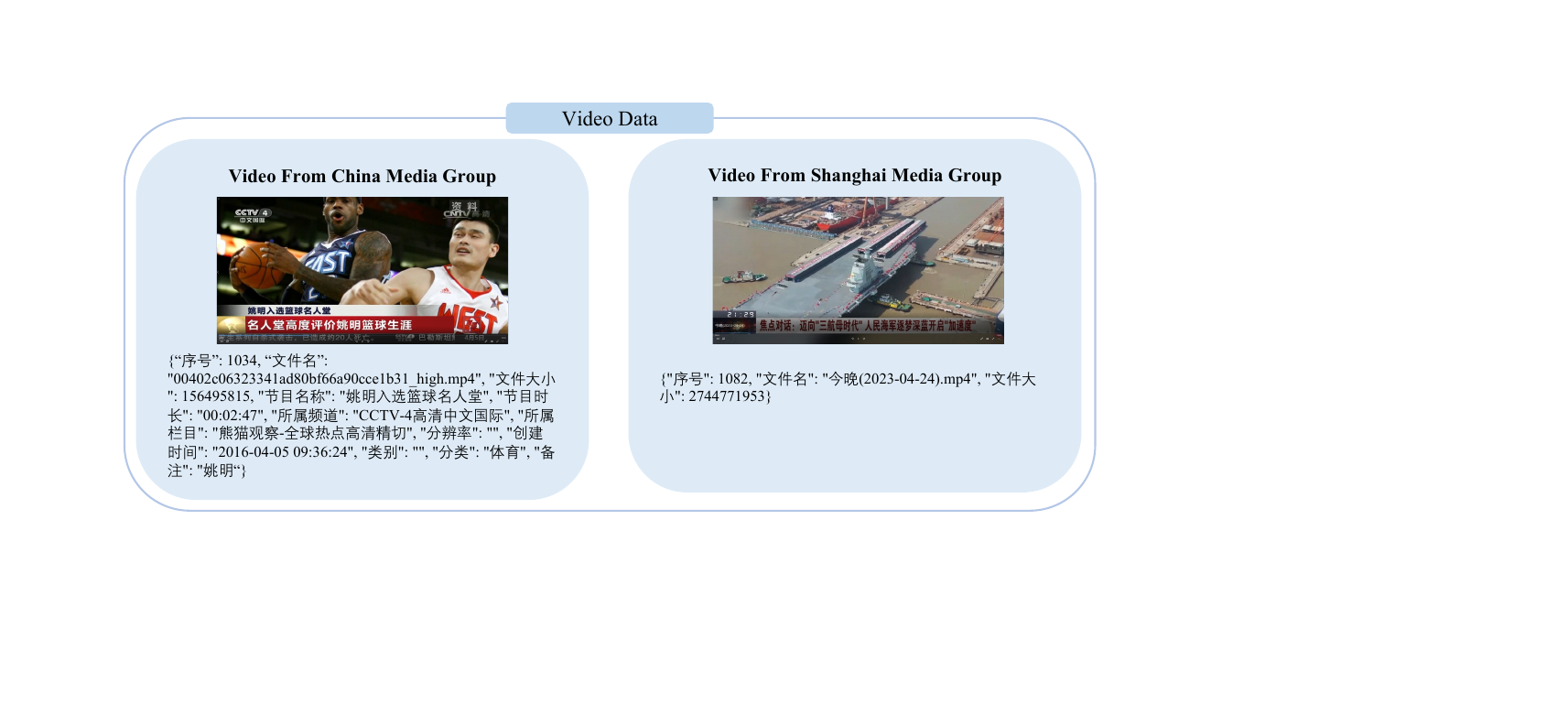}
\end{center}
		\caption{Examples of Video Data}
\label{fig:fig3-video}
\end{figure*}

The video data is sourced from the high quality of the China Media Group and Shanghai Media Group program footage. It encompasses over 1000 videos, with a data size surpassing 900GB. Figure~\ref{fig:fig3-video} illustrates examples of the video data. The inclusion of video data allows for the exploration of tasks that require understanding and generating content across different modalities, such as video captioning and video question answering.


\begin{table*}[htb]
\centering
\begin{tabular}{llccc}
\toprule
Data Type & Source & Weight(\%) & Number of Files & Size(GB) \\
\midrule
 Video Data     & CMG \& SMG & 100 & 1K  & 916.7\\ 
\bottomrule
\end{tabular}
\caption{Video Data Statistics from China Media Group (CMG) and Shanghai Media Group (SMG).}
\label{tab:video_data}
\end{table*}

\section{Methods}

The video data, sourced from the China Media Group (CMG) and Shanghai Media Group (SMG), is of high quality. Thus, we mainly focused on cleaning text and text-image data, and this section introduces the data collection, cleaning, and value alignment process.

\subsection{Text Data Cleaning}
To ensure the comprehensiveness of the data, our text corpus combines data from eight different sources, as fully outlined in Table 2. For the English internet data provided by the Common Crawl, we employed a multi-step text extraction process, language detection, corpus filtering, and deduplication to obtain high-quality data.

Firstly, we extracted text from the original WARC files, then used different language detection tools (pycld2) to classify the extracted text, subsequently processing the Chinese and English texts differently. Given the abundance of invalid data on the internet, we then applied the following rules to filter and obtain high-quality data:
\begin{itemize}
    \item We removed irregular documents, including those with inappropriate average word and document lengths. If the most frequent word was non-alphabetic, or the frequency was too high, we considered it an uncommon document format and deleted it.

    \item We removed documents with too little content, such as those with fewer than three sentences after processing; fewer than three paragraphs; fewer than three paragraphs longer than 200 words; or fewer than two stopwords.

    \item We also cleaned paragraphs, removing special sections such as those containing words like JavaScript, sections outside of punctuation marks, and paragraphs with more than 1000 words. Even a small amount of duplicate data can significantly impact our observations. We noticed that the obtained data contained duplicates, so we tokenized the text data and used MinHashLSH and n-grams to evaluate similarity, deleting content with a similarity greater than 0.8.
\end{itemize}

Furthermore, due to the presence of harmful and low-quality content in internet data, we trained some models to assess quality and filter for different issues:
\begin{itemize}
    \item We trained content safety models for pornography, violence, gambling, attacks, and other toxic themes using the FastText model, separately for Chinese and English, to filter out potentially toxic data.

    \item We trained data quality models for various low-quality data found online, such as auto-generated random data and advertising content, separately for Chinese and English, to reduce the proportion of low-quality data.
\end{itemize}

Based on the above filtering, we obtained safe, high-quality, value-aligned text data.

\subsection{Image-Text Data Cleaning}

The interleaved image-text data comes from four sources, as detailed in Table 2. Since the text-image data come from official sources, they are of high quality. Here, we selectively extracted the needed content to form the interleaved image-text data. For the formation of interleaved image-text data, we followed these steps:
\begin{itemize}
    \item To reduce the difficulty of cleaning and ensure data quality, we wrote specific parsing rules for each site. User-generated articles were sourced from a single open-source site.
    
    \item We only extracted valid (ad-free, list-free, navigation bar-free, emoticon-free, comment-free) article content. We used a series of rules to filter. We also removed references, complex tables, lists, and other entry-related content for the text part of Wikipedia, retaining only the text paragraphs. For user-generated articles and authoritative media news, we used XPath, CSS selectors, and regular expressions to remove media sources, publishers, reposts, advertisements, and comments unrelated to the article theme, to obtain the article's main body. Like with text data cleaning, we also performed deduplication based on similarity.
    
    \item We believed the header images of Wikipedia articles are meaningful for image selection, so we only retained these. To ensure all images in user-generated articles had valid descriptions, we removed articles with more than 15 images and those where the number of text characters was less than twice the number of images. Based on this rule, we retained 55\% of the valid articles.

    \item We  the valid text and images obtained from the above filtering. The format for Wikipedia was (the first paragraph, main image, and remaining paragraphs).
\end{itemize}

After applying these filters, we obtained high-quality interleaved text-image data. The language distribution was 62.3\% Chinese and 37.7\% English.

\section{Conclusion}

In this paper, we introduced the \textbf{WanJuan} dataset, a large-scale, multimodal Chinese-English dataset collected from diverse web sources. The dataset includes various modalities, such as text, image-text, and video, all meticulously processed to ensure safety, richness of content, and accuracy. This high-quality dataset provides a valuable resource for the training of large language models and the study of various multimodal tasks. We believe that the release of this dataset will significantly contribute to the advancement of research in the fields of Natural Language Processing and Computer Vision, especially for tasks that require understanding and generating content across different modalities.

\newpage

\bibliographystyle{plain}
\bibliography{wanjuan.bib}

\end{document}